\documentclass[sigconf]{acmart}

\usepackage{algorithm}
\usepackage{algorithmic}
\usepackage{diagbox}
\usepackage{multirow}
\usepackage{subcaption}
\usepackage{amsfonts}
\usepackage{amsmath}
\usepackage[switch]{lineno}
\usepackage{balance}
\usepackage{flushend}
\AtBeginDocument{%
  \providecommand\BibTeX{{%
    \normalfont B\kern-0.5em{\scshape i\kern-0.25em b}\kern-0.8em\TeX}}}

\copyrightyear{2021} 
\acmYear{2021} 
\setcopyright{acmlicensed}
\acmConference[CIKM '21]{Proceedings of the 30th ACM International Conference on Information and Knowledge Management}{November 1--5, 2021}{Virtual Event, QLD, Australia}
\acmBooktitle{Proceedings of the 30th ACM International Conference on Information and Knowledge Management (CIKM '21), November 1--5, 2021, Virtual Event, QLD, Australia}
\acmPrice{15.00}
\acmDOI{10.1145/3459637.3482052}
\acmISBN{978-1-4503-8446-9/21/11}
\begin{document}
\fancyhead{}
%%
%% The "title" command has an optional parameter,
%% allowing the author to define a "short title" to be used in page headers.
\title{Adversarial Learning for Incentive Optimization in Mobile Payment Marketing}

%%
%% The "author" command and its associated commands are used to define
%% the authors and their affiliations.
%% Of note is the shared affiliation of the first two authors, and the
%% "authornote" and "authornotemark" commands
%% used to denote shared contribution to the research.
%\author{Xuanying Chen*, Zhining Liu*, Yu Li*, Sen Li, Lihong Gu, Xiaodong Zeng, Yize Tan, Jinjie Gu}
%\affiliation{
%\institution{Ant Financial Services Group}
%\thanks{*These authors contributed equally to this work.}
%%\city{Hangzhou}
%%\country{China}
%}
%\email{{xuanying.cxy, eason.lzn, jinli.yl, lisen.lisen, lihong.glh, xiaodong.zxd, yize.tyz, jinjie.gujj}@antgroup.com}

\author{Xuanying Chen*}
\affiliation{
\institution{Ant Financial Services Group}
}
\thanks{*These authors contributed equally to this work.}
\email{xuanying.cxy@antgroup.com}

\author{Zhining Liu*}
\affiliation{
\institution{Ant Financial Services Group}
}
\email{eason.lzn@antgroup.com}

\author{Li Yu*}
\affiliation{
\institution{Ant Financial Services Group}
}
\email{jinli.yl@antgroup.com}

\author{Sen Li}
\affiliation{
\institution{Alibaba Group}
}
\email{lisen.lisen@alibaba-inc.com}

\author{Lihong Gu}
\affiliation{
\institution{Ant Financial Services Group}
}
\email{lihong.glh@antgroup.com}

\author{Xiaodong Zeng}
\affiliation{
\institution{Ant Financial Services Group}
}
\email{xiaodong.zxd@antgroup.com}

\author{Yize Tan}
\affiliation{
\institution{Ant Financial Services Group}
}
\email{yize.tyz@antgroup.com}

\author{Jinjie Gu}
\affiliation{
\institution{Ant Financial Services Group}
}
\email{jinjie.gujj@antgroup.com}

%%
%% By default, the full list of authors will be used in the page
%% headers. Often, this list is too long, and will overlap
%% other information printed in the page headers. This command allows
%% the author to define a more concise list
%% of authors' names for this purpose.
\renewcommand{\shortauthors}{Chen and Liu, et al.}

%%
%% The abstract is a short summary of the work to be presented in the
%% article.
\begin{abstract}
Many payment platforms hold large-scale marketing campaigns, which allocate incentives to encourage users to pay through their applications.
To maximize the return on investment, incentive allocations are commonly solved in a two-stage procedure. After training a response estimation model to estimate the users' mobile payment probabilities (MPP), a linear programming process is applied to obtain the optimal incentive allocation. However, the large amount of biased data in the training set, generated by the previous biased allocation policy, causes a biased estimation. This bias deteriorates the performance of the response model and misleads the linear programming process, dramatically degrading the performance of the resulting allocation policy. To overcome this obstacle, we propose a bias correction adversarial network. Our method leverages the small set of unbiased data obtained under a full-randomized allocation policy to train an unbiased model and then uses it to reduce the bias with adversarial learning. 
Offline and online experimental results demonstrate that our method outperforms state-of-the-art approaches and significantly improves the performance of the resulting allocation policy in a real-world marketing campaign.
\end{abstract}

%%
%% The code below is generated by the tool at http://dl.acm.org/ccs.cfm.
%% Please copy and paste the code instead of the example below.
%%


%%
%% Keywords. The author(s) should pick words that accurately describe
%% the work being presented. Separate the keywords with commas.

%%
%% This command processes the author and affiliation and title
%% information and builds the first part of the formatted document.
\maketitle

\section{Introduction}
Mobile payments such as 
Alipay, WeChat Pay, Apple Pay are now experiencing rapid growth.
To maximize the return on investment, how to allocate the user-specific incentive under budget constraints is playing the central role in marketing budget allocation.

~\citeauthor{chu2013study}~\cite{chu2013study} ~\cite{gupta2008allocating} divided the budget allocation into two stages: 
adapting a response model to estimate the user’s response score, i.e., MPP in our paper,
and applying a linear programming to obtain the optimal incentive allocation under budget constraints.
In this paper, we will focus on the first step to learn a better response model and not elaborate the whole literature of the second step.

\begin{figure}[t]
    \centering
    \begin{subfigure}[b]{0.2\textwidth}
        \includegraphics[width=\textwidth,height=0.8\textwidth]{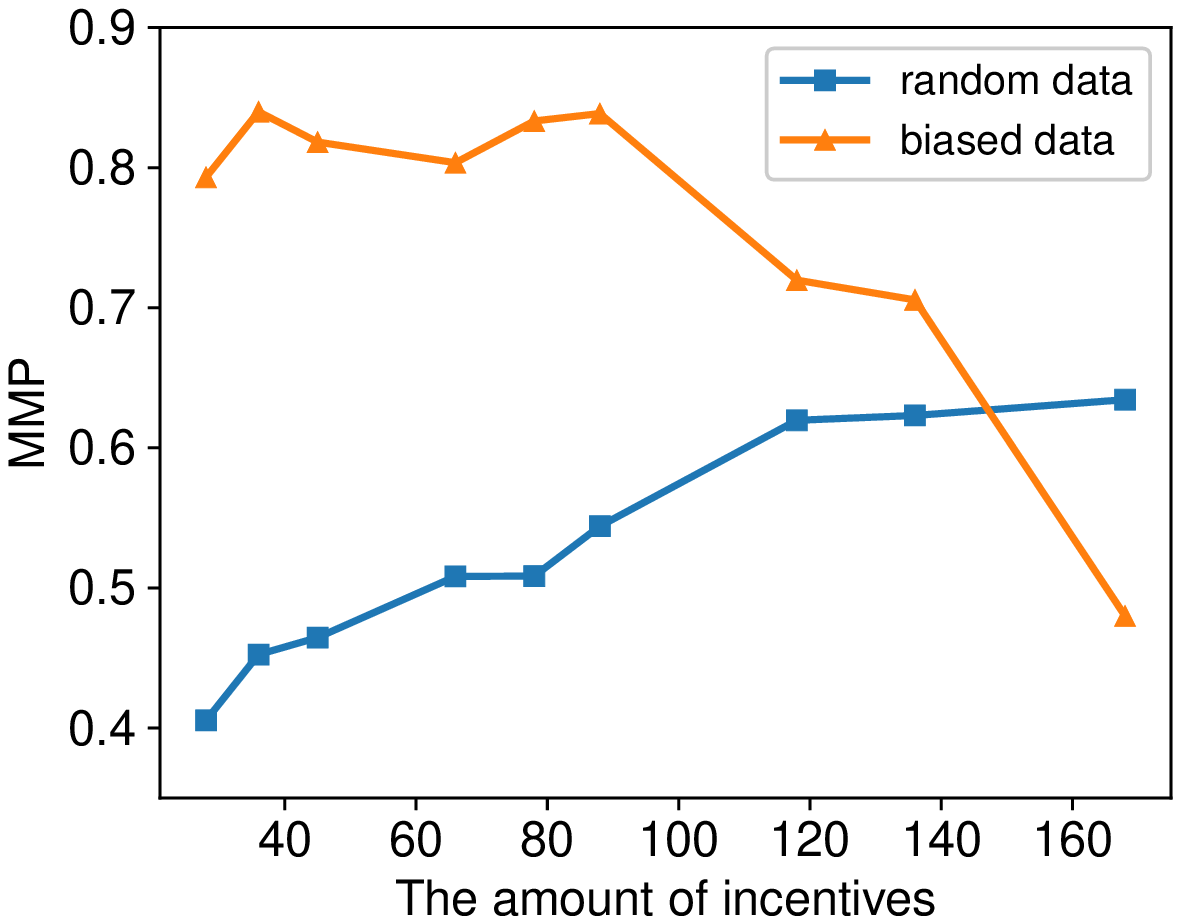}
        \caption{}
        \label{fig:gull}
    \end{subfigure}%
    ~ %add desired spacing between images, e. 
    \begin{subfigure}[b]{0.2\textwidth}
        \includegraphics[width=\textwidth,height=0.8\textwidth]{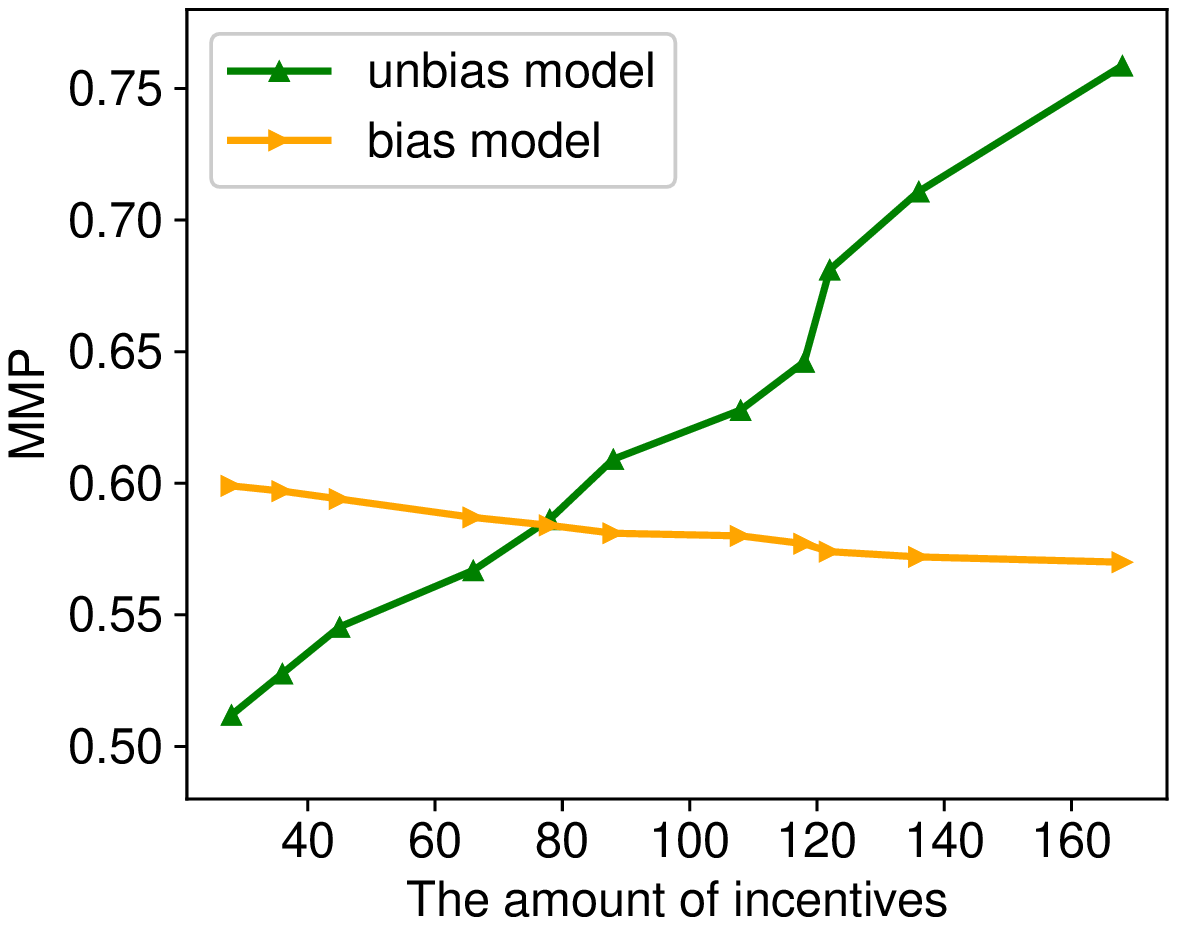}
        \caption{}
        \label{fig:tiger}
    \end{subfigure}
    \caption{The curve of the mobile payment probability and the scale of incentives on payment data. (a) Pure random data versus previous biased data; (b) Response model trained with random data versus biased data.}\label{fig:animals}
    \vspace{-2mm}
\end{figure}

It is non-trivial to estimate the response score of each user to incentives, since the training payment data is typically generated by previous biased allocation policy.
Figure~\ref{fig:gull} shows the relationship between MMP and the amount of incentives under random data and biased data. 
Under random data, the MPP increases as the incentive increases, which describes real user behavior. 
On the contrary, the previous bias allocation policy generates a large amount of biased data, e.g., active users have a high probability of being allocated a small amount of incentive, while low-active users are the opposite.
As a result, the MPP-incentives curve shows a downward trend.
Specifically, due to the naturally high MPP of active users, even if a small amount of incentives are allocated, they are more likely to pay through the applications. 
On the contrary, even if a large amount of incentives are allocated to low-active users, their interest in paying through the applications is also low.
The response model trained with these data will over-estimate the response score on a small amount of the incentive and under-estimate the response score on a large amount of the incentive, shown in Figure~\ref{fig:tiger}, which cannot accurately estimate the user's response score. As a result, the linear programming algorithm will tend to allocate a relative small amount of incentive to users.
One solution to get unbiased estimation is to adapt a full-randomized allocation policy~\cite{bonner2018causal} to collect a large amount of unbiased data, which is impractical due to the limited budget. 
In this paper, we propose a price-bias correction adversarial network (PCAN), which leverages the small set of unbiased data obtained under a full-randomized allocation policy to train an unbiased network and then uses the unbiased network to reduce the price-bias in the biased network with adversarial learning. 
Specifically, PCAN first learns to distinguish the distribution difference between the biased and unbiased data representation and then teaches the biased network to generate a representation close to the unbiased network, which can alleviate the problem of price-bias.

Our paper is organized as follows. Related works are reviewed in Section 2, followed by our proposed method in Section 3. Experimental results are reported in Section 4 before we conclude the paper in the last section.

\section{Related Work}
Existing incentives allocation methods usually are divided into two stages: (1) the response model estimates the response score; (2) the response score is served as an input of the optimization model, which tries to maximize the MPP under the budget constraint.

% 这里似乎还包含了一些跟allocation无关的paper，感觉没有必要去讨论
\noindent
\textbf{Response Model Estimation.} \cite{akter2016big,10.5555/2804919.2804920} introduce a dynamic marketing allocation budget. Through strong fitting ability, neural networks can achieve a high prediction accuracy, widely used in many scenarios~\cite{liao2018deep,taieb2017regularization,tong2017simpler}. \cite{ferreira2016analytics,ye2018customized} use machine learning techniques to estimate future demand for new products and provide recommended prices for Airbnb hosts. However, as a black-box model, there are some gaps between the prediction and decision-making of deep neural networks~\cite{PMID:28154050}. Therefore, ~\cite{zhao2019unified,liu2019graph}~propose a semi-black box model that extends the logarithmic demand curve through neural network and graph learning to solve this problem.

\noindent
\textbf{Allocation Optimization.}
\cite{ito2017optimization} proposes a fast approximation using semi-definite programming relaxation. \cite{ferreira2016analytics} optimizes pricing decisions by using demand predictions from the regression trees as inputs of the price optimization model. 
\cite{boutilier2016budget,staib2017robust} through markov decision process value function and the connection with continuous sub-module function to solve the allocation optimization problem.

% \begin{figure*}[t]
% 	\centering
%     \includegraphics[width=1.0\textwidth,height=0.25\textwidth,trim=40 210 140 160, clip]{model.eps}
% 	\caption{An illustration of the proposed framework.}
% 	\label{fig:framework}
% \end{figure*}
\begin{figure*}[t]
	\centering
    \includegraphics[scale=0.45,trim=50 200 100 150, clip]{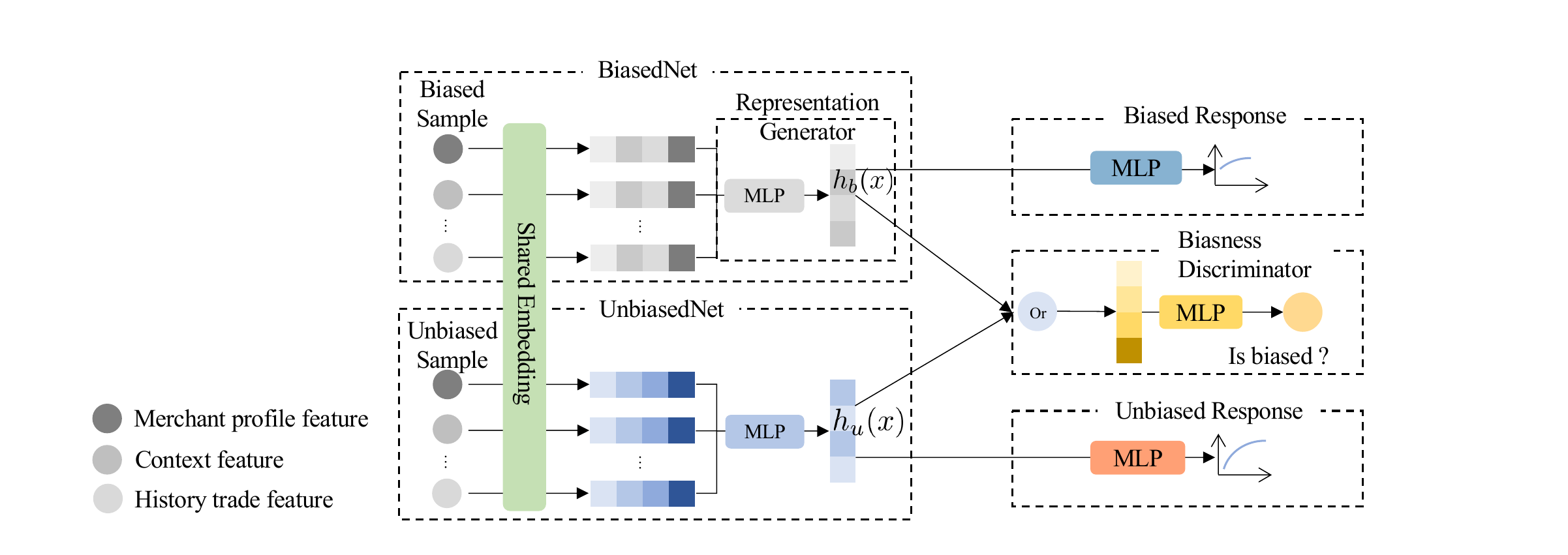}
	\caption{An illustration of the proposed framework.}
	\label{fig:framework}
\end{figure*}
 %%%%%%%%%%%%%%%%%%%%缩减竖直距离%%%%%%%%%%%%%%%%%%%%%%
\section{Proposed Method}

From the online deployed system, we can collect the logged data $D=(X, T, Y)$, where $x\in X$ denotes a feature vector that encodes the information of users' demographic profiles and online behaviors. $t\in T$ represents the amount of incentives allocated to the users, which are usually several preset numbers. $y \in Y$ is the label, which denotes whether users make mobile payments.

To allocate the budget in an optimal way, a two-stage solution~\cite{liu2019graph,zhao2019unified} is a common choice. First, a response model $y=f(x,t)$ estimates the users’ MPPs for each user described by $x$ and each price $t$. 
Based on the response model, the best allocation under the constraint of the budget $B$ can be obtained by solving a linear programming problem. In the following subsections, we illustrate our proposed method from the two perspectives in detail.

\subsection{Response Model Estimation}
With the collected data $D$, the response model $f$ can be estimated through minimizing the following loss function:
\begin{equation}
\setlength{\abovedisplayskip}{3pt}
\setlength{\belowdisplayskip}{3pt}
	\mathcal{L} = \sum_{(y,x,t)\in D}CrossEntropy(y ,f(x, t))
	\label{eq:pred_loss}
\end{equation}
where $f$ is usually defined as:
\begin{equation}
\setlength{\abovedisplayskip}{3pt}
\setlength{\belowdisplayskip}{3pt}
	f(x_i, t_i) = \sigma(w_g(h(x_i)) \cdot t_i + w_p(h(x_i)))
	\label{eq:response_model}
\end{equation}
where $h(x_i)$ denotes the latent representation of the user. $\sigma(\cdot)$ denotes the sigmoid function. $w_g$ and $w_p$ are two trainable functions. To further meet the monotonicity constraint, a semi-black-box model with Softplus (SBBM-Softplus)~\cite{liu2019graph} is introduced to guarantee the positive effect of the treatment:
\begin{equation}
\setlength{\abovedisplayskip}{3pt}
\setlength{\belowdisplayskip}{3pt}
	f(x_i, t_i) = \sigma (\text{SOFTPLUS}[w_g(h(x_i))] \cdot t_i + w_p(h(x_i)))
	\label{eq:mono_mapping}
\end{equation}
where $\text{SOFTPLUS}(x)=\ln (1+e^x)$.

\subsection{Price-Bias Correction Adversarial Network}
However, due to the budget constraints, we would not deploy a random allocation policy online. Training based on logged data from a biased allocation policy,$f(x_i, t_i)$ is easily biased. As pointed by \cite{schnabel2016recommendations,yuan2019improving}, the observed accurate response score is estimated using Inverse Propensity Scoring (IPS)-based methods:
\begin{equation}
\setlength{\abovedisplayskip}{3pt}
\setlength{\belowdisplayskip}{3pt}
	f_{true}(x_i,t_i) = \frac{f(x_i,t_i)}{\pi(t_i|x_i)}
	\label{eq:mono_mapping}
\end{equation}
where $\pi(t_i|x_i)$ denotes the probability for the user $x_i$ to be allocated with the incentive $t_i$.

But IPS-based estimators cannot handle well significant shifts in exposure probability between
treatment and control policies
under biased exposure.
Therefore, to correct this bias, in our learning setup, we assume that we have access to a large sample $D_b$ from the online biased allocation policy and a small sample $D_u$ from the randomized allocation policy, i.e., $D=D_b\cup D_u$. Specifically, as shown in Figure~\ref{fig:framework}, our entire model includes two subnets named BiasedNet $f_b$ and UnbiasedNet $f_u$. Two subnets are trained to optimize $\mathcal{L}_b$ and $\mathcal{L}_u$, respectively, where $\mathcal{L}_b$ and $\mathcal{L}_u$ are defined as:
\begin{equation}
\begin{aligned}
\setlength{\abovedisplayskip}{3pt}
\setlength{\belowdisplayskip}{3pt}
	\mathcal{L}_b = \sum_{(y,x,t)\in D_b}CrossEntropy(y ,f(x,t))\\
	\mathcal{L}_u = \sum_{(y,x,t)\in D_u}CrossEntropy(y ,f(x,t))
\end{aligned}
\label{eq:double-tower}
\end{equation}

Although the distributions of $D_b$ and $D_u$ are different, we assume that the same set of users should have a fixed distribution representation. 
Therefore, we would like to enforce $h_b(x)$ to follow the same distribution of $h_u(x)$ to alleviate the bias. Specifically, we realize the goal of debiasing using adversarial learning through optimize the following objective function:
\begin{equation}
\begin{split} 
\setlength{\abovedisplayskip}{3pt}
\setlength{\belowdisplayskip}{3pt}
\min_{h_b} \max_{d} \mathop{\mathbb{E}}_{x\sim D_u}[\log d(h_u(x))]+\mathop{\mathbb{E}}_{x\sim D_b}[\log (1-d(h_b(x))]
\label{eq:g_loss}
\end{split}
\end{equation}

As shown in the equation, two components play the central role of representation generator and bias discriminator. We describe two parts in detail in the following subsections.

\subsubsection{Representation Generator} BiasedNet, which also acts as the representation generator to generate a biased latent representation in the training process of the adversarial network. In contrast, a user described by $h_u(x)$ in UnbiasedNet represents an unbiased latent representation, which serves as the supervision of $h_b(x)$. The representation generator will align $h_b$ to $h_u$ to eliminate deviation.

\subsubsection{Bias Discriminator} We introduce the bias discriminator $d$ to determine which dataset it comes from, i.e., $D_b$ and $D_u$. Specifically, one batch of training data is mixed with unbiased/biased samples, and $d$ is trained to maximize the probability of correctly identifying which dataset $h_b(x)$ is generated from. In an adversarial way, $h_b$ is trained to maximize the probability of $d(x)$ incorrectly identifying. 
Since $h_u$ is trained by unbiased data, the learned representation $h_u(x)$ is assumed to be unbiased. Under the guidance of the adversarial network, BiasNet can gradually learn the distribution of $h_u(x)$, and finally achieve bias correction.

\begin{algorithm}[t]
	\caption{Training Process of PCAN}
	\label{Alg1}
	\begin{algorithmic}[1]
		\renewcommand{\algorithmicrequire}{ \textbf{Input:}}     %Use Input in the format of Algorithm
		\renewcommand{\algorithmicensure}{ \textbf{Output:}}    %UseOutput in the format of Algorithm
		\REQUIRE ~~\\
		Unbiased dataset $D_u$, biased dataset $D_b$ and warm-up step $n_w$
		\ENSURE ~~\\
		Response Model $f^*$\\
		\STATE Random initialization of the model\\
		\STATE\ \textit{step}$\ \leftarrow \ 0$ \\
		\WHILE{not converged}
		\WHILE{\textit{step} $<= n_w$}
		\STATE Sample one batch of data from $D_u$ and $D_b$, respectively
		\STATE Update BiasedNet and UnbiasedNet by optimizing Eq.\ref{eq:double-tower}
		\STATE\ \textit{step} $\ \leftarrow \ $ \textit{step}$\ +\ 1$\\
		\ENDWHILE\\
        \WHILE{\textit{step} $<= n_w+6$}
		\STATE Sample one batch of data from $D_u$ and $D_b$, respectively
		\STATE Update alternately BiasedNet and $d$ by optimizing Eq.~\ref{eq:g_loss}
		\STATE\ \textit{step} $\ \leftarrow \ $ \textit{step}$\ +\ 1$\\
		\ENDWHILE
		\STATE\ \textit{step}$\ \leftarrow \ 0$ \\
		\ENDWHILE
		\STATE Return BiasedNet.
	\end{algorithmic}
\end{algorithm}

% With the help of the representation generator and the bias discriminator, the representations of biased users $h_b(x)$ can be forced to be similar to the representations of unbiased users $h_u(x)$  so that Eq.~\ref{eq:g_loss} and Eq.~\ref{eq:d_loss} can be optimized. The optimal solution is that $h_u(x)$ and $h_b(x)$ to achieve the same distribution, which indicates that the bias problem is alleviated.

\subsubsection{Optimization Algorithm} To train the response model, we need to alternately optimize Eq.~\ref{eq:double-tower} and Eq.~\ref{eq:g_loss}, which requires carefully update the discriminator and generator. Algorithm~\ref{Alg1} summarizes the training details. The algorithm includes two phases of training. The first phase is the warm-up~\cite{you2017large}, which is designed to ensure the training of adversarial components to start from a relatively good situation. Here the warm-up step $n_w$ is set to 100 in this paper. Then in the second phase, we begin to update the representation generator and bias discriminator under the framework of adversarial learning. Moreover, to enhance the stability of training~\cite{goodfellow2014generative}, we alternate between five steps of optimizing $d(x)$ and one step of optimizing $h_g(x)$. The algorithm finally returns the learned response model $f^*$ after convergence.

\subsection{Allocation Optimization as Linear Programming}
Based on the estimated response model, we can get the mobile payment probability $f(x_i, t_j)$ for each user with the short form $f_{ij}$.
Assuming that $T$ is the treatment list, which is defined as $T=(t_1,...,t_{|T|})$.
Then we formalize the allocation as a linear programming problem given the payment probability $f_{ij}$ and the budget $B$, and the objective is to maximize the sum of mobile payment probability over the user set $M$:
\begin{equation}
\setlength{\abovedisplayskip}{3pt}
\setlength{\belowdisplayskip}{3pt}
\begin{aligned}
&\max \sum_{i=1}^{M} k_i \sum_{j=1}^{|T|} f_{ij} * a_{ij}\\%,\, \text{ s.t. }\quad 
%s.t. \quad & a_{ij} \in [0, 1], for \quad i=1,...,M, j=1,...,|T|  \\
%& \sum_{j=1}^{|T|} a_{ij} = 1, for \quad i=1,...,M \\
s.t.\ &\frac{\sum_{i=1}^{M} k_i \sum_{j=1}^{|T|} a_{ij} * t _j}{\sum_{i=1}^{M} k_i} <= B
\end{aligned}
\end{equation}
where $k_i$ is the number of incentives.
$a_{ij}$ is the indicator of whether choosing the incentive $t_j$ for user $x_i$.
%Under constraints $\sum_{j=1}^{|T|} a_{ij} = 1$, $\sum_{j=1}^{|T|} a_{ij} * t_j$ is the value of incentive decided for each user.
%We then solve the convex optimization problem with the Lagrange multiplier method.
The optimal solution is :
\begin{equation}
\begin{aligned}
\setlength{\abovedisplayskip}{3pt}
\setlength{\belowdisplayskip}{3pt}
arg\min \limits_{j} f_{ij} - \lambda  (t_j - B) \quad for \quad j=1,...,|T| \\
\end{aligned}
\end{equation}
where $\lambda$ denote the dual optimal.

\section{EXPERIMENTS}
In this section, we conduct offline and online experiments on the proposed method to demonstrate its effectiveness. Before diving into experimental results, we first introduce experiment settings.
\subsection{Experimental Settings}
In this section, we first introduce our dataset for training the model and experimental settings.

\subsubsection{Dataset} 
We collected two experimental datasets separately from two real-world mobile marketing campaigns.\\
\textbf{Dataset A}: Contains more than 50 million samples which include 11 kinds of incentives. 5\% of the data is collected by a random strategy. The remaining 95\% is collected by the biased strategy.\\
\textbf{Dataset B}: Contains millions of samples which include 16 kinds of incentives, and the sample ratio is the same as dataset A.\\
\textbf{Test Set}: All the test set in this paper is the fully random dataset.
%Random strategy: Only a few users are selected and assigned a random amount of incentives, and we name it the unbiased data $D_u$.
%We can assume that those users with different sensitivities to incentives are randomly sampled treatments.
%Online serving model: The rest of the data is collected from the online serving model named biased data $D_b$, where the incentive allocation is laughed to find the best strategy to maximize payment probability.
%There are more than 1.6 million of labeled users are observed whether to pay

%\noindent
% \subsubsection{Comparison Methods} We compare our method with several strong baselines: \textbf{SBBM-U}~\cite{zhao2019unified}, \textbf{SBBM-B}~\cite{zhao2019unified}, \textbf{SBBM-Softplus}~\cite{liu2019graph}, inverse probability weighting (\textbf{IPW})~\cite{rosenbaum1983central} and \textbf{CausE}~\cite{bonner2018causal}.
\subsubsection{Comparison Methods} We compare our method with several baselines:
\begin{itemize}
	\item \textbf{SBBM-U}~\cite{zhao2019unified}: A baseline model training based on $D_u$.
	\item \textbf{SBBM-B}~\cite{zhao2019unified}: Similar with SBBM-U, which train the response model on the dataset $D=D_u\cup D_b$.
	\item \textbf{SBBM-Sp}~\cite{liu2019graph}: A response model with the constraint of the monotonicity trained based on $D=D_u\cup D_b$.
	\item \textbf{IPW}~\cite{rosenbaum1983central}: The method which weights each sample with the inverse of its propensity score.
%	 to achieve unbiased estimator training on $D=D_u\cup D_b$.
	\item \textbf{CausE}~\cite{bonner2018causal}: The approach uses the data from $D=D_u\cup D_b$, introducing a regularizer to minimize the difference between the weights of the two models.
%	to better predict randomized treatment effects on pairs of users and products.
\end{itemize}

\vspace{0.5pt}
\noindent
\subsubsection{Metrics} We introduce Area-Under-Curve (AUC) to quantitatively measure the performance of different methods. Besides, price calibration error (PCE) is also introduced according to expected calibration error (ECE)~\cite{guo2017calibration}, which is defined as:
\begin{equation}
	\text{PCE} = \sum_{i=1}^{M} \frac{1}{M}|\mathop{\mathbb{E}}\limits_{i\in real}[y_i] - \mathop{\mathbb{E}}\limits_{i\in pred}[y_i]|
\end{equation}
where $M$ represents the number of amounts of the incentive.
PCE is designed to measure the difference between the average prediction scores and labels.
Due to the confidentiality of data, all metrics are presented the relative improvement over the baseline.

Table~\ref{table:cold_eval} shows the experiment result on the dataset.
We can observe that PCAN performs the best on the two metrics overall baseline methods across all the two datasets. To further demonstrate the effectiveness of the proposed method in correcting the bias, we show average response scores over a group of users with randomly allocated amount of incentives which shows the average estimated MPP of different models for different amounts in Figure~\ref{fig:off1}. 
%Except for SBBM-U, all other models are trained using $D_u\cup D_b$. 
Comparing these models trained on the biased dataset, PCAN successfully captures the monotonicity. Besides, PCAN is the most similar to the average score of an actual label in most cases.

\subsection{Offline Results}
\begin{table}[h]
\renewcommand\tabcolsep{2pt} % 调整表格列间的长度
\small
	\centering
	\caption{Performance comparison of different methods over SBBM-U.}
	\label{table:cold_eval}
	\begin{tabular}{@{}llcccccc@{}}
		\toprule
	   & \multicolumn{7}{c}{\textbf{Model}} \\
		& \textbf{Data}     & SBBM-U     & SBBM-B     & SBBM-Sp    & IPW    & CausE &PCAN         \\ \midrule
		\multirow{2}{*}{AUC}     & \textbf{A}    &-    &  -83.16\%     & +5.67\%            & +3.87\%        & +18.32\%      & \textbf{+61.92\%}     \\   
		& \textbf{B}        &  -     & -1.54\%      & +7.74\%        & +6.16\%   & +8.76\%    & \textbf{+9.78\%}    \\ \midrule
	\multirow{2}{*}{PCE}     & \textbf{A}    &-    &  -38.38\%     & +36.58\%            & +41.80\%        & +39.30\%      & \textbf{+44.35\%}     \\   
		& \textbf{B}        &  -     & -9.58\%      & +15.01\%        & +14.12\%   & +17.36\%    & \textbf{+19.04\%}    \\ \midrule
	\end{tabular}
	\vspace{-1em}%%%%%%%%%%%%%%%%%%%%缩减竖直距离%%%%%%%%%%%%%%%%%%%%%%
\end{table}

\begin{figure}
    \centering
    \begin{subfigure}[b]{0.21\textwidth}
        \includegraphics[width=\textwidth]{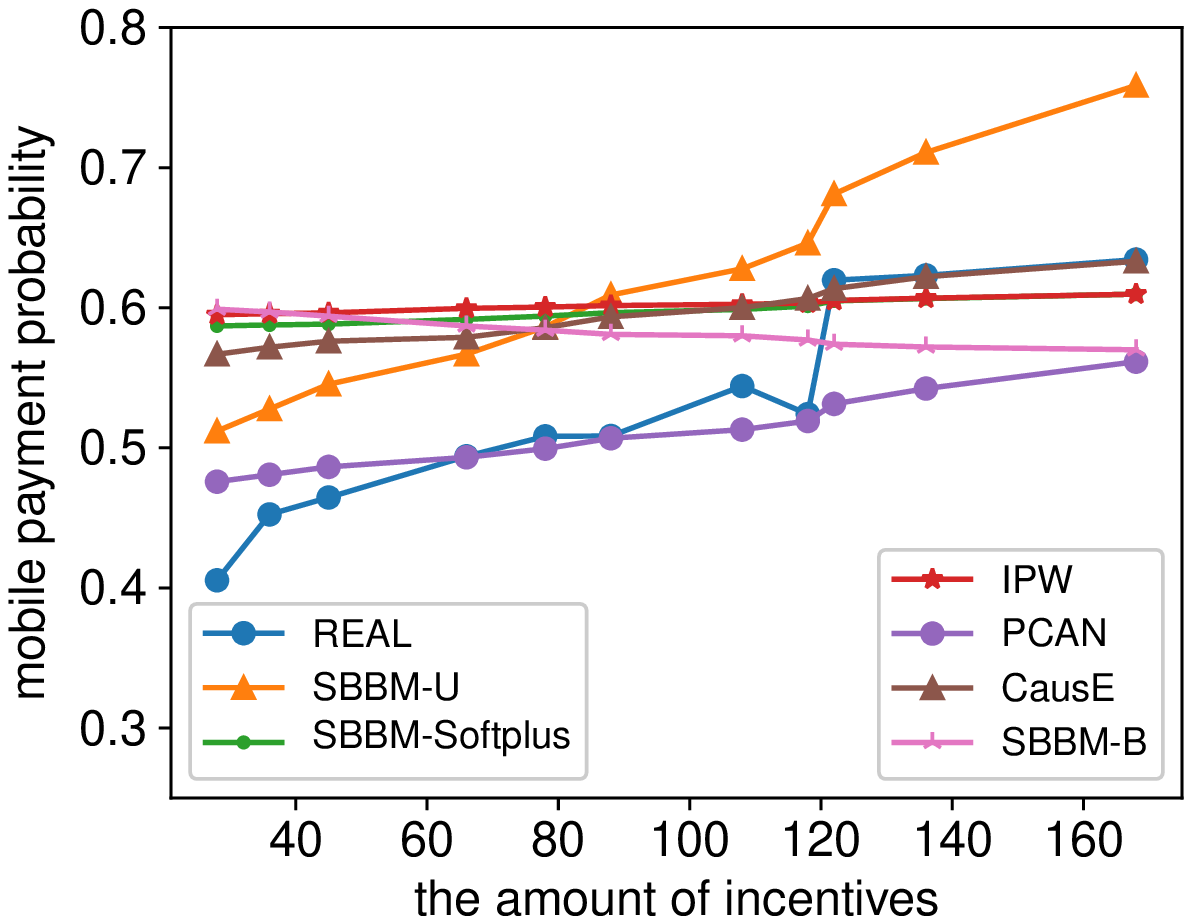}
        \caption{}
        \label{fig:off1}
    \end{subfigure}%
    ~ %add desired spacing between images, e. g. ~, \quad, \qquad etc.
      %(or a blank line to force the subfigure onto a new line)
    \begin{subfigure}[b]{0.21\textwidth}
        \includegraphics[width=\textwidth]{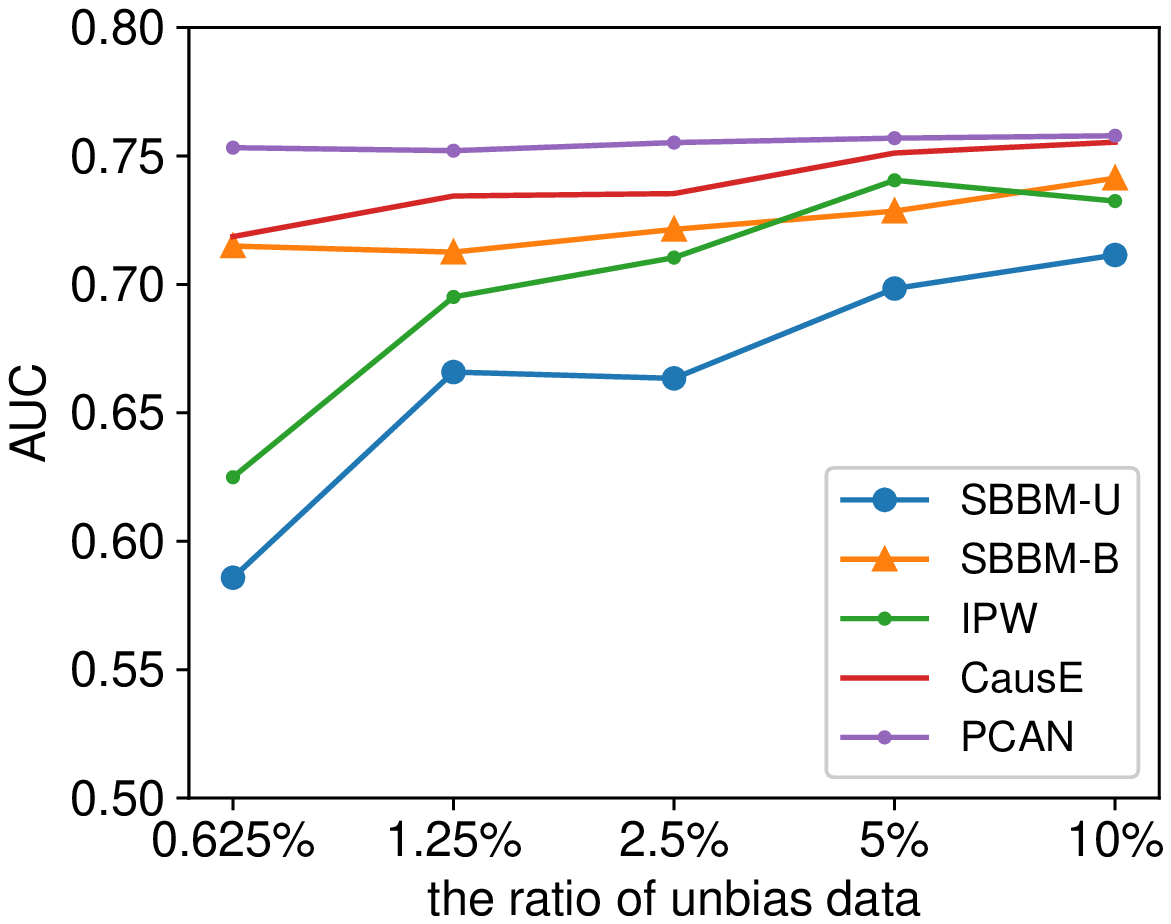}
        \caption{}
        \label{fig:off2}
    \end{subfigure}
    \caption{(a) The response curve of different methods; (b) The performance of difference methods over different ratio of random data.}
    \label{fig:animals}
    \vspace{0.1cm}
\end{figure}

In addition, all the above results are tested with the ratio of $|D_u|:|D_b|=19:1$. To prove the robustness of our model, we change the proportion of $|D_u|$ to view the performance of each model. It can be seen in Figure~\ref{fig:off2} that 1) As the proportion of $|D_u|$ increases, the performance of all models gradually improves. 2) As $|D_u|$ shrinks, the performance of all methods gradually deteriorates. However, our proposed framework, PCAN, still performs better than other frameworks. In particular, for other methods, the AUC value is strongly affected by the size of $|D_u|$, but our method is still stable. Since the cost of an unbiased sample $|D_u|$ is relatively high in actual marketing activities, PCAN can achieve good results at a relatively low $|D_u|$ sample ratio, save a lot of costs, and has strong robustness.
 
\subsection{Online Results}
To verify the proposed method's effectiveness, we further conducted an A/B test against a baseline (\textbf{SBBM-Softplus}). In the A/B test, we first randomly split all candidates into two buckets. Under the same budget, the baseline and our approach allocate the incentives to users in the two buckets based on estimated response models, respectively. Over two marketing campaigns, we both observe an over 3\% increase in the number of mobile payments.

\section{Conclusion}
In this paper, we propose an adversarial learning method for incentive optimization in mobile payment marketing. We identify the bias in the response model estimated from the biased data by analyzing the response-incentive curve. We further introduce the mechanism of adversarial learning to build an unbiased response model. Comparing with other state-of-the-arts methods, The online experiment results verify that our proposed method can significantly increase mobile payment usage under a limited budget.
Future work will design a more effective method with approximate the accurate response model with less randomized allocation data.

\bibliographystyle{ACM-Reference-Format}
% \balance
% \bibliography{reference.bib}

%%% -*-BibTeX-*-
%%% Do NOT edit. File created by BibTeX with style
%%% ACM-Reference-Format-Journals [18-Jan-2012].

\end{document}